\documentclass[arxiv]{melba}

\usepackage{amsmath,amsfonts}
\usepackage{booktabs}
\usepackage{array}
\usepackage{graphicx}
\usepackage{enumitem}

\melbaid{YYYY:NNN}
\doi{10.59275/j.melba.YYYY-AAAA}
\melbaauthors{Forootan et al.}
\email{$^*$ hamidreza@databiox.com}
\volume{X}
\firstpageno{1}
\melbayear{2026}
\datesubmitted{2025-6-1}
\datepublished{2025-6-1}

\ShortHeadings{GIM-ENDO Dataset}{Forootan et al.}

\title{GIM-ENDO: A Multimodal Endoscopic Image and Video Dataset\\
for Gastric Intestinal Metaplasia Morphology and Pathology}

\author{
\firstname Mojgan \surname Forootan\aff{1},
\firstname Mahziar \surname Setayeshfar\aff{2},
\firstname Ali \surname Darvishi\aff{3},
\firstname Mohammad \surname Tashakoripour\aff{4},
\firstname Hamidreza \surname Bolhasani\aff{5,*}
}

\affiliations{%
\num 1 \addr Shahid Beheshti University of Medical Sciences, Tehran, Iran \\
\num 2 \addr Iran University of Medical Sciences, Tehran, Iran \\
\num 3 \addr Shiraz University of Medical Sciences, Shiraz, Iran \\
\num 4 \addr Tehran University of Medical Sciences, Tehran, Iran \\
\num 5 \addr DataBioX Research, Tehran, Iran
}

\abstract{%
Gastric intestinal metaplasia (GIM) is a precursor lesion to gastric dysplasia and
adenocarcinoma whose early detection is crucial for intervening in the carcinogenesis
cascade. Artificial intelligence (AI) holds considerable promise for real-time endoscopic
detection and characterization of GIM. However, development of reliable AI models has
been constrained by the absence of publicly available, histopathologically validated
datasets that combine detailed endoscopic annotations, histological subtype (complete
and incomplete), standardized grading systems, and normal mucosal patterns. GIM-ENDO
was designed to fill this gap. The dataset comprises demographic data, endoscopic
findings, histopathological results, and \textit{H.~pylori} status acquired using the
Olympus EVIS X1 system with white-light endoscopy (WLE) and image-enhanced endoscopy
(IEE), including narrow-band imaging (NBI) and magnifying NBI (M-NBI), along with
images and video clips from 24 patients (22 GIM-positive, 2 normal controls).
Annotations cover six primary IEE endoscopic signs --- light blue crest (LBC), marginal
turbid band (MTB), white opaque substance (WOS), TV pattern (Fusion), atrophy, and
map-like erythema (MLE) --- plus two additional endoscopic findings (AHP and GA)
recorded where present. GIM subtypes (complete and incomplete) are annotated for all
GIM-positive cases; OLGA and OLGIM staging are provided where complete histological
sampling was available. The dataset is publicly accessible at
\url{https://doi.org/10.5281/zenodo.20707267}. For the latest updates and further
website: \url{https://databiox.com}.
\\[12pt]\textit{A short version of this work has been accepted at the MICCAI 2026 Open Data Track.}}

\begin{document}
\twocolumn[\maketitle]

\section{Background}

Gastric cancer is the fifth most common malignancy and the fourth leading cause of
cancer death globally, accounting for approximately 1.1~million new cases and 769,000
deaths each year~\citep{sung2021}. Most gastric adenocarcinomas develop through the
Correa~\citep{correa1992} cascade --- normal mucosa $\to$ chronic gastritis $\to$
atrophic gastritis $\to$ intestinal metaplasia $\to$ dysplasia $\to$ carcinoma ---
making early detection at the intestinal metaplasia stage a cornerstone of cancer
prevention. GIM is regarded as the first irreversible step in this progression, and its
presence substantially elevates cancer risk, particularly in the incomplete
subtype~\citep{gonzalez2020,wang2023}.

AI-assisted analysis of IEE images has emerged as a promising complement to endoscopic
practice. Deep learning systems match or exceed expert accuracy for detecting
gastrointestinal lesions~\citep{hirasawa2018,mori2018}, and initial GIM-specific studies
have shown that deep learning models can detect and characterize metaplastic areas in
IEE images with encouraging accuracy~\citep{dohi2020,martins2025}. Progress toward
clinical translation remains bottlenecked by the scarcity of openly accessible,
annotated, multimodal datasets that combine histopathological validation with IEE-sign
annotations and normal mucosal controls.

\subsection{Related Resources}

To our knowledge, few publicly available datasets integrate histopathologically
validated GIM cases with detailed IEE-sign annotations, standardized histological
grading, and normal mucosal controls within a single resource.

\cite{yang2023} released 21,493 WLE and linked-colour-imaging images for binary
IM/atrophy detection, without IEE-sign characterization or standardized histological
grading. GastroHUN~\citep{bravo2025} provides 8,834 images and 4,729 videos across 22
anatomical landmarks but lacks GIM-specific annotations. HyperKvasir~\citep{borgli2020}
covers a broad range of gastrointestinal findings without dedicated GIM labeling. A
comparative summary of these resources is provided in Table~\ref{tab:comparison}.

\begin{table*}[t]
\centering
\caption{Comparative summary of existing GIM-related datasets and AI studies.}
\label{tab:comparison}
\small
\setlength{\tabcolsep}{5pt}
\begin{tabular}{lccccc}
\toprule
\textbf{Dataset} & \textbf{Pub.} & \textbf{N} & \textbf{Sub.} &
\textbf{IEE} & \textbf{OLGA} \\
\midrule
Yang et al.~\citep{yang2023}         & \checkmark & 21k    & --         & --         & -- \\
Pornvoraphat et al.~\citep{pornvoraphat2024} & --  & --     & --         & Segm.      & -- \\
Ligato et al.~\citep{ligato2024}     & \checkmark & 1,384  & --         & --         & -- \\
Iwaya et al.~\citep{iwaya2023}       & --         & --     & --         & --         & -- \\
Fang et al.~\citep{fang2024}         & --         & 2,725  & \checkmark & --         & \checkmark \\
Martins et al.~\citep{martins2025}   & --         & --     & --         & --         & -- \\
GastroHUN~\citep{bravo2025}          & \checkmark & 8,834  & --         & --         & -- \\
HyperKvasir~\citep{borgli2020}       & \checkmark & 110k+  & --         & --         & -- \\
\textbf{GIM-ENDO (Ours)}             & \checkmark & 99     & \checkmark & \checkmark & \checkmark \\
\bottomrule
\end{tabular}

\smallskip\noindent\footnotesize
Pub.: publicly available; N: images; Sub.: GIM subtype annotation;
IEE: IEE-sign annotations; OLGA: OLGA/OLGIM staging; Segm.: segmentation masks only.
\end{table*}

\subsection{Novelty}

The main novelty of this research lies in its potential to assist gastroenterologists in
improving targeted biopsies of precancerous lesions --- specifically GIM and atrophy ---
based on endoscopic morphological features such as LBC, MTB, WOS, TV pattern with
fusion, TM, and MLE. Furthermore, it may contribute to the detection of dysplasia at
earlier stages in the future.

Such early detection could facilitate timely endoscopic resection, thereby interrupting
the progression cascade at an early point. However, this remains a hypothesis.
Prospective studies with larger sample sizes and long-term follow-up are required to
validate this approach. If confirmed, this strategy could represent a turning point in
gastrointestinal oncology.

\section{Summary}

\subsection{Vision and Objectives}

Our goal is to develop an AI model based on endoscopic findings for the early detection
of gastric premalignant lesions. To this end, we aim to identify and extract key
endoscopic features --- including WOS, LBC, MTB, the TV pattern, atrophy, MLE, and
other relevant markers --- and to classify lesions into complete versus incomplete
subtypes.

By leveraging these features, we seek to train a robust and reliable AI model
deployable to gastroenterologists worldwide. Such a model could improve the targeting
of biopsy sites and support decision-making for endoscopic resection with appropriate
peripheral margins, ultimately enhancing early detection and effective management of
these lesions.

\subsection{AI Readiness}

Data are provided in standard image and video formats (JPEG, PNG, MP4) with consistent
file naming conventions. All associated metadata and annotations are stored in a
structured Microsoft Excel (\texttt{.xlsx}) file, which includes per-image and per-case
labels such as annotated endoscopic features, GIM subtype, OLGA/OLGIM stage, and
relevant clinical information (e.g., \textit{H.~pylori} status).

All images and video clips are organized within a single directory structure, with
linkage to the metadata file established through unique identifiers. No predefined
train/validation/test split is provided, allowing researchers to define task-specific
data partitioning strategies based on their experimental needs.

\subsection{Resources Needed}

The dataset is publicly accessible at \url{https://doi.org/10.5281/zenodo.20707267}.
Standard Python libraries (e.g., NumPy, Pandas, OpenCV, PyTorch, or TensorFlow) are
sufficient for data loading, preprocessing, and model development.

Given the relatively small size of the current dataset ($\approx$98 images and 39 video
clips of $\sim$50 seconds each), experiments can be conducted on standard CPU-based
systems. GPU use is optional but may accelerate deep learning training. For the latest
updates and dataset expansions, readers are encouraged to visit
\url{https://databiox.com/}.

\section{Methods}

\subsection{Data Details}

Consecutive adults ($\geq$18 years) undergoing upper gastrointestinal endoscopy for
dyspepsia, screening, or surveillance indications were prospectively enrolled across
multiple centers. Exclusion criteria included prior gastric surgery, a history of
gastric malignancy, active upper gastrointestinal bleeding, and inadequate biopsy
samples for histological assessment. Demographic and pathological characteristics are
summarized in Table~\ref{tab:demographics}.

\begin{table}[t]
\centering
\caption{Demographic and pathological characteristics of the study population.}
\label{tab:demographics}
\small
\begin{tabular}{lcc}
\toprule
\textbf{Characteristic} & \textbf{GIM+} & \textbf{Control} \\
\midrule
$N$ (patients)                       & 22          & 2 \\
Images / Videos                      & 92 / 34     & 6 / 5 \\
Age, Median (IQR)$^{\dagger}$        & 59 (55--68) & N/A \\
Sex, Female $N$ (\%)                 & 14 (63\%)$^{\dagger}$ & 1 (50\%) \\
Sex, Male $N$ (\%)                   & 8 (36\%)$^{\dagger}$  & 1 (50\%) \\
\textit{H.~pylori} +, $N$ (\%)      & 4 (20\%)$^{\dagger}$  & --- \\
Gastropathy, $N$ (\%)                & 6 (27\%)    & 0 \\
Complete GIM, $N$                    & 8           & --- \\
Incomplete GIM, $N$                  & 7           & --- \\
Complete + Incomplete, $N$           & 3           & --- \\
Subtype Unconfirmed, $N$             & 4$^{\ddagger}$ & --- \\
OLGA Stage I--II, $N$                & 5           & --- \\
OLGA Stage III--IV, $N$              & 1           & --- \\
OLGA Not Assessed, $N$               & 16$^{\S}$   & --- \\
\bottomrule
\end{tabular}

\smallskip\noindent\footnotesize
$^{\dagger}$ Age/sex missing for 1 patient (2617) pending clinical record retrieval.\\
$^{\ddagger}$ GIM-positive, subtype not further classified: Patients 2609, 2617, 2618.\\
$^{\S}$ OLGA not assessed in 15/21 cases due to incomplete biopsy-site sampling.
\end{table}

The distribution of key endoscopic signs (LBC, MTB, WOS, TV pattern, TM, and MLE)
among GIM-positive cases at the patient level is presented in
Table~\ref{tab:signs}.

\begin{table}[t]
\centering
\caption{Distribution of endoscopic signs in GIM-positive cases (patient level).}
\label{tab:signs}
\small
\begin{tabular}{lc}
\toprule
\textbf{Sign} & \textbf{GIM+ ($N = 22$)} \\
\midrule
LBC (Light Blue Crest)        & 16 / 22 (73\%) \\
MTB (Marginal Turbid Band)    & 5 / 22 (23\%) \\
WOS (White Opaque Substance)  & 4 / 22 (18\%) \\
TV pattern (Fusion)           & 12 / 22 (55\%) \\
TM (Transparent Mucosa)       & 14 / 22 (64\%) \\
MLE (Map-like Erythema)       & 5 / 22 (23\%) \\
\bottomrule
\end{tabular}

\smallskip\noindent\footnotesize
\textit{A patient is counted positive for a sign if at least one image shows that sign.}
\end{table}

\subsection{Methods Used for Data Creation}

All procedures were performed using the Olympus EVIS X1 system (processor: CV-1500;
Olympus Corp., Tokyo, Japan). In most cases, a CF-EZ1500DL gastroscope was used;
however, in a subset of examinations, a 190-series gastroscope (CF-H190L) was utilized.
Imaging modalities included white-light endoscopy (WLE), magnifying narrow-band imaging
(M-NBI), and narrow-band imaging (NBI) with near focus.

\begin{figure}[t]
\centering
\includegraphics[width=\linewidth]{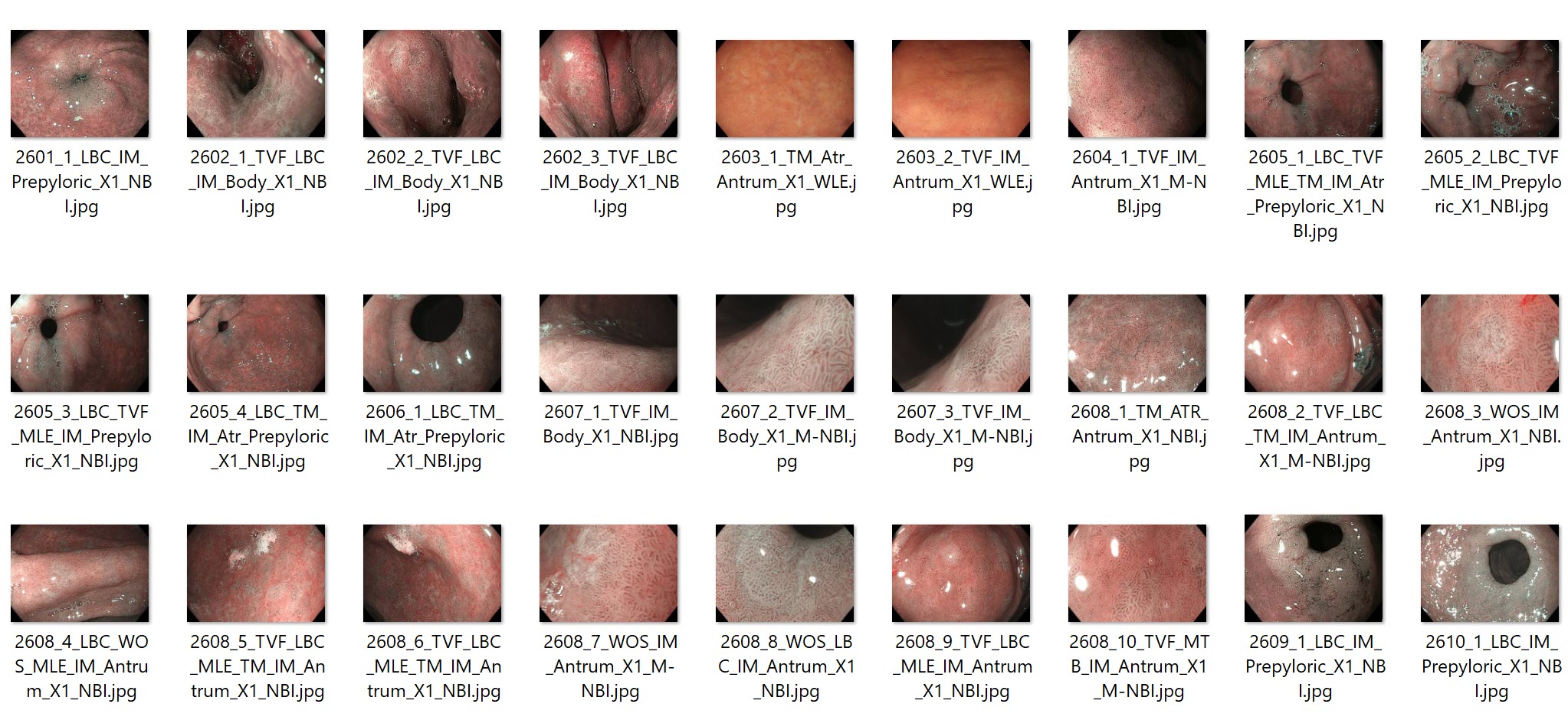}
\caption{Representative endoscopic images from the GIM-ENDO dataset. The file naming
convention encodes key annotations, including anatomical location, imaging modality,
and detected endoscopic features (e.g., LBC, WOS, TV pattern), facilitating structured
labeling and data traceability.}
\label{fig:endoscopic}
\end{figure}

Still images were captured in JPEG and PNG formats at $2879\!\times\!1799$~pixels
(standard full-frame) and $657\!\times\!454$~pixels (magnified near-focus view). Video
clips (MP4, 30~fps, approximately 50~seconds, $1920\!\times\!1080$ resolution) were
obtained from relevant anatomical locations for each patient (antrum, body/corpus,
incisura angularis, and prepyloric region). Each location was documented as both a
still image and a corresponding video clip. Representative examples are shown in
Figure~\ref{fig:endoscopic}.

\begin{figure}[t]
\centering
\includegraphics[width=\linewidth]{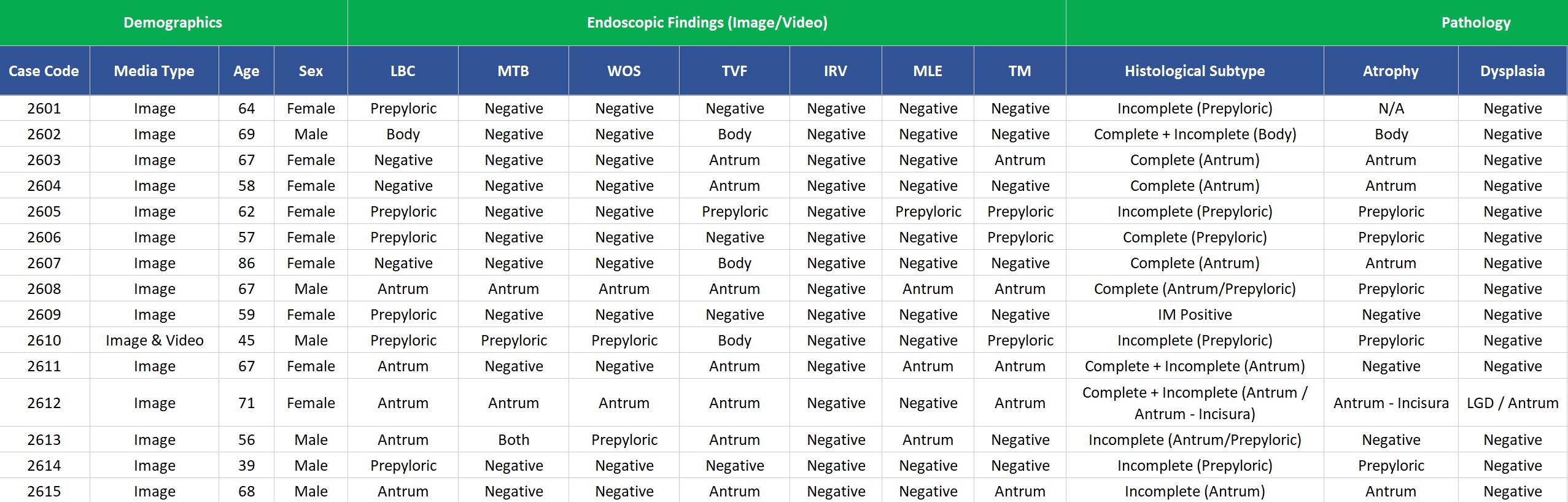}
\caption{Example of the structured annotation and metadata file (XLSX) used in the
GIM-ENDO dataset. The table includes demographic information, expert-annotated
endoscopic findings (e.g., LBC, MTB, WOS, TV pattern, MLE, TM), and corresponding
histopathological outcomes, including GIM subtype, atrophy, and dysplasia.}
\label{fig:metadata}
\end{figure}

In the initial release, a total of 137 samples were included, comprising 98 still
images and 39 video clips. All cases were confirmed by histopathological examination.
Endoscopic findings included WOS, LBC, MTB, the TV pattern, atrophy, MLE, and other
relevant features. Following identification of suspicious lesions, targeted biopsies
were obtained and sent for histopathological evaluation. Upon pathological confirmation,
the corresponding cases were recorded in a structured Excel database as metadata. The
associated still images and video clips were then selected and annotated by experienced
gastroenterologists (advanced endoscopists) according to predefined criteria. A
representative excerpt of the annotation file is shown in Figure~\ref{fig:metadata}.

\section{Validation}

\subsection{Quality Control and Ground-Truth Verification}

All image and video files were inspected for completeness, correct anatomical labeling,
and absence of identifiable information prior to inclusion. Ground truth was established
through a four-step process:
\begin{enumerate}[leftmargin=*,topsep=2pt,itemsep=1pt]
  \item Targeted biopsies obtained according to the updated Sydney Protocol~\citep{dixon1996};
  \item Independent histopathological assessment by two board-certified gastrointestinal pathologists;
  \item Expert annotation of endoscopic findings by multiple experienced gastroenterologists (advanced endoscopists);
  \item Consensus-based resolution of any discordant cases.
\end{enumerate}

Cases without complete biopsy-site coverage were excluded from definitive OLGA staging
and were appropriately flagged in the annotation file.

\subsection{Annotation Consistency}

Endoscopic-sign annotation was performed by multiple experienced gastroenterologists
(advanced endoscopists), with standardized definitions applied consistently throughout
the dataset. Formal inter-rater and intra-rater reliability were not assessed in this
pilot release; this constitutes a known limitation acknowledged in
Section~\ref{sec:limitations}. Users should note that certain features, particularly
MTB and TV-pattern subcategories, are inherently subject to inter-observer variability.

\subsection{Benchmark Protocol}

No predefined train/validation/test split or benchmarking protocol is included in the
current release. Users are encouraged to define task-specific data partitioning
strategies (e.g., patient-level splits) based on their experimental design.

The dataset supports several potential tasks, including binary GIM detection
(GIM-positive vs.\ normal), subtype classification (complete vs.\ incomplete), per-sign
detection of endoscopic features, and OLGA/OLGIM staging. Evaluation metrics and
operating points (e.g., AUROC, sensitivity at fixed specificity) should be selected
according to the intended clinical application.

Given the current dataset size, this resource is well suited for exploratory analyses,
pilot studies, and proof-of-concept model development. Ongoing efforts are focused on
expanding the dataset with additional cases, images, and video data, which will enable
more comprehensive benchmarking and robust model validation in future releases.

\section{Discussion}

\subsection{Strengths}

In this study, we present GIM-ENDO as a clinically grounded and systematically
annotated dataset designed to accelerate reproducible AI research in gastric
premalignant lesions. By integrating high-quality endoscopic imaging,
histopathological confirmation, and structured metadata, we aimed to leverage AI to
reduce the gap between the interpretation of endoscopic findings and histopathological
confirmation, and in doing so, to facilitate early detection and enable early resection
with free margins, ultimately contributing to reduced mortality and morbidity associated
with gastric cancers.

A key strength of this study lies in the use of multimodal endoscopic data, including
both still images and video clips acquired under standardized imaging modalities (WLE,
M-NBI, and NBI with near focus). Unlike many prior studies that rely solely on static
images, the inclusion of video data provides a more comprehensive and dynamic
representation of mucosal patterns, which may improve the robustness and
generalizability of AI models. Furthermore, all cases were confirmed by
histopathological examination, ensuring high diagnostic validity and minimizing
labeling errors.

Another important feature of this dataset is the detailed annotation of endoscopic
findings --- including WOS, LBC, MTB, the TV pattern, atrophy, and MLE --- performed
by experienced gastroenterologists (advanced endoscopists). This level of granularity
enables not only binary detection of GIM, but also supports more advanced tasks such
as per-feature detection and subtype classification (complete vs.\ incomplete). Such
an approach is essential for developing interpretable AI systems that align with
clinical decision-making in endoscopy.

Importantly, the study design reflects real-world clinical workflows: suspicious
endoscopic findings prompted targeted biopsies, followed by histopathological
confirmation and structured data recording. This linkage between endoscopy and
pathology enhances the clinical relevance of the dataset and supports the development
of AI models aimed at improving biopsy targeting and early detection. Moreover, this
integration may facilitate more accurate risk stratification and staging, including the
potential automation of OLGA/OLGIM systems.

Despite these strengths, several limitations should be acknowledged. First, the dataset
size in its initial release (137 samples, including 98 still images and 39 video clips)
remains relatively modest for training deep learning models, particularly for more
complex tasks. Second, annotations were performed by a single expert, which may
introduce observer bias; future studies should incorporate multi-expert consensus to
enhance reliability. Third, although data acquisition was performed using a consistent
processor platform, variability in endoscope types and real-world imaging conditions
may still influence model performance.

Future work should focus on expanding the dataset with larger and more diverse cohorts,
incorporating multi-center data, and conducting prospective validation studies. In
addition, leveraging temporal information from video sequences and integrating clinical
variables may further improve model performance. Ultimately, the deployment of such AI
systems in clinical practice has the potential to support gastroenterologists in
optimizing biopsy targeting, improving resection strategies with appropriate margins,
and enabling earlier detection of gastric premalignant lesions.

In conclusion, GIM-ENDO represents an important step toward clinically meaningful AI
applications in gastrointestinal endoscopy. With further development and validation,
it has the potential to contribute to a paradigm shift in the early detection and
management of gastric premalignant disease.

\subsection{Limitations and Biases}
\label{sec:limitations}

The primary limitation of this study is the relatively small sample size, reflecting
the early stage of dataset development. While all cases were carefully curated with
histopathological confirmation and expert annotation, larger cohorts will be essential
to further improve model robustness and generalizability.

In addition, the number of normal control cases is currently limited. This is partly
attributable to the high prevalence of \textit{Helicobacter pylori} infection and
environmental risk factors in the studied population, which reduce the availability of
truly normal gastric mucosa in routine clinical practice. As a result, the dataset is
enriched for pathological findings, which should be considered when developing and
calibrating downstream AI models.

Future work will focus on expanding the dataset with a larger and more balanced cohort,
particularly by increasing the number of cases and, consequently, the volume of
corresponding images and video data. This expansion is expected to mitigate current
limitations and further enhance the robustness, generalizability, and clinical
applicability of the proposed AI framework.

\subsection{Responsible Use}

GIM-ENDO is intended for non-commercial research on gastric premalignant disease,
supporting tasks such as binary GIM detection, subtype classification, per-sign
detection, and OLGA/OLGIM staging. Users must not attempt to re-identify any
participant.

Models developed using this dataset should not be deployed in clinical settings without
independent prospective validation in the target population and geographic context. The
dataset reflects a symptomatic or surveillance-referred population from multiple
centers; therefore, prevalence rates and feature distributions may differ from those in
unselected screening populations, and model recalibration may be required before
application in other settings.

The underrepresentation of rare OLGA/OLGIM stages (III--IV) and incomplete GIM
subtypes should be acknowledged in any downstream analysis or publication.
Re-identification attempts, commercial use, and use in surveillance or decision-making
applications are strictly prohibited under the terms of the CC BY-NC 4.0 license
(\url{https://creativecommons.org/licenses/by-nc/4.0/}).

\section{Resource Availability}

\subsection{Data Location}

GIM-ENDO is publicly available at \url{https://doi.org/10.5281/zenodo.20707267}. The
repository contains endoscopic still images (JPEG and PNG, 98 files), video clips (MP4
and MKV, 39 files), and anonymised clinical metadata, including \textit{H.~pylori}
status. All annotations and metadata are provided in a structured Excel (\texttt{.xlsx})
file. All files are organized with consistent naming conventions, and linkage between
media files and metadata is established via unique identifiers. No predefined
train/validation/test split is included in the current release. Access is open and
unrestricted. For the latest updates, visit \url{https://databiox.com}.

\subsection{Use Cases and Licensing}

Target ML tasks: binary GIM detection, subtype classification (complete
vs.\ incomplete), per-sign detection (LBC, MTB, WOS, TV pattern, atrophy, MLE, AHP,
GA), and OLGA/OLGIM staging. Clinical domain: upper GI endoscopy for pre-malignant
lesion screening and risk stratification. The dataset is released under CC BY-NC 4.0
(\url{https://creativecommons.org/licenses/by-nc/4.0/}) for non-commercial research use.

\acks{Not applicable.}

\ethics{This study has been approved by the ethics committee of the Gastroenterology
and Liver Disease Research Center, Research Institute for Gastroenterology and Liver
Diseases, Shahid Beheshti University of Medical Sciences (Ethics Approval No.:
IR.SBMU.RIGLD.REC.1405.039). According to ethical
principles, the datasets are completely anonymous. Informed consent was obtained from
all subjects and/or their legal guardian(s).}

\coi{The authors declare no competing financial or non-financial interests.}

\data{GIM-ENDO is openly available at \url{https://doi.org/10.5281/zenodo.20707267}.
The repository contains endoscopic still images (JPEG/PNG), video clips (MP4/MKV), and
anonymised clinical metadata (XLSX) including \textit{H.~pylori} status and all
IEE-sign annotations. Access is open and unrestricted.}

\bibliography{gimendo_full}

\clearpage
\section*{Author Biographies}

\noindent\begin{minipage}[t]{0.22\linewidth}
\vspace{0pt}
\includegraphics[width=\linewidth]{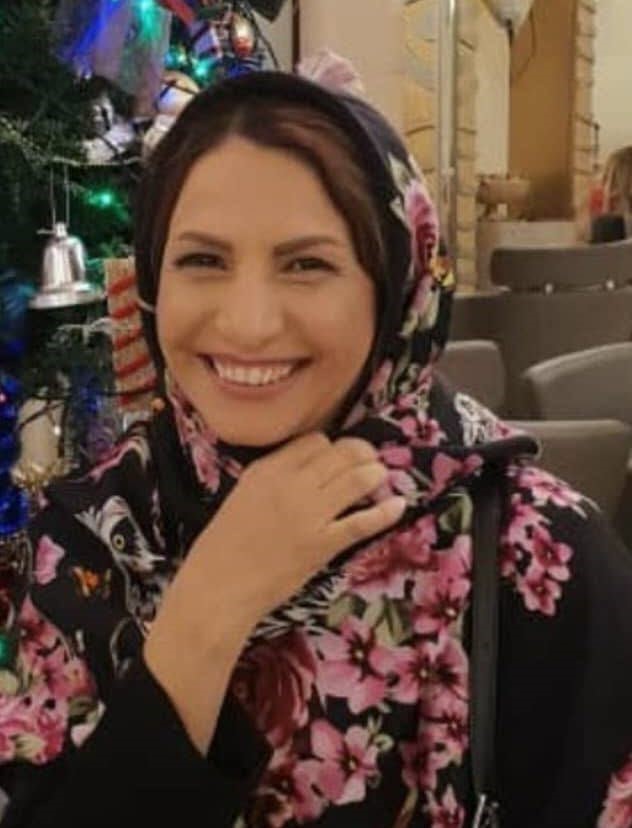}
\end{minipage}
\hfill
\begin{minipage}[t]{0.74\linewidth}
\vspace{0pt}
\textbf{Mojgan Forootan, MD} is a Professor of Gastroenterology at
Shahid Beheshti University of Medical Sciences, Tehran, Iran. Her research focuses on gastrointestinal oncology, endoscopic diagnosis
of precancerous lesions, and AI-assisted endoscopy.
\end{minipage}

\bigskip
\noindent\begin{minipage}[t]{0.22\linewidth}
\vspace{0pt}
\includegraphics[width=\linewidth]{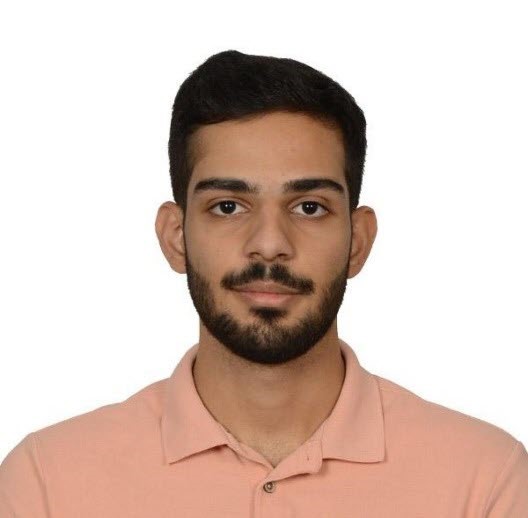}
\end{minipage}
\hfill
\begin{minipage}[t]{0.74\linewidth}
\vspace{0pt}
\textbf{Mahziar Setayeshfar} is a Medical Student at Iran University of
Medical Sciences, Tehran, Iran. His research interests include medical data science,
clinical AI applications, and gastrointestinal imaging.
\end{minipage}

\bigskip
\noindent\begin{minipage}[t]{0.22\linewidth}
\vspace{0pt}
\includegraphics[width=\linewidth]{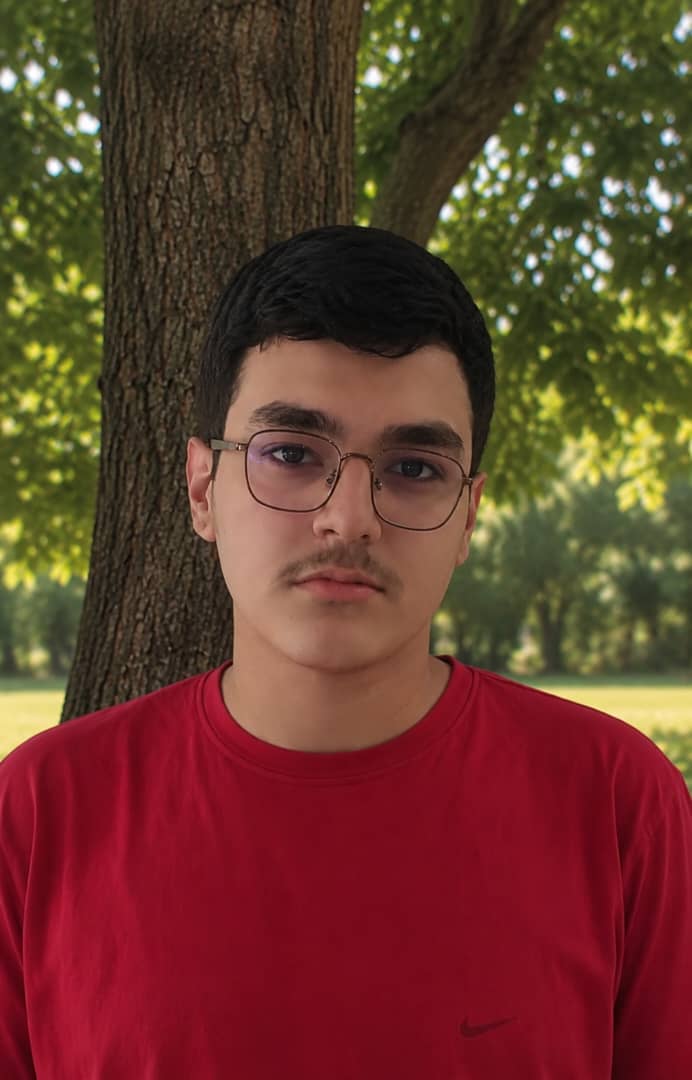}
\end{minipage}
\hfill
\begin{minipage}[t]{0.74\linewidth}
\vspace{0pt}
\textbf{Ali Darvishi} is a Medical Student at Shiraz University of
Medical Sciences, Shiraz, Iran. His research interests include endoscopic imaging,
data curation, and AI applications in gastroenterology.
\end{minipage}

\bigskip
\noindent\begin{minipage}[t]{0.22\linewidth}
\vspace{0pt}
\includegraphics[width=\linewidth]{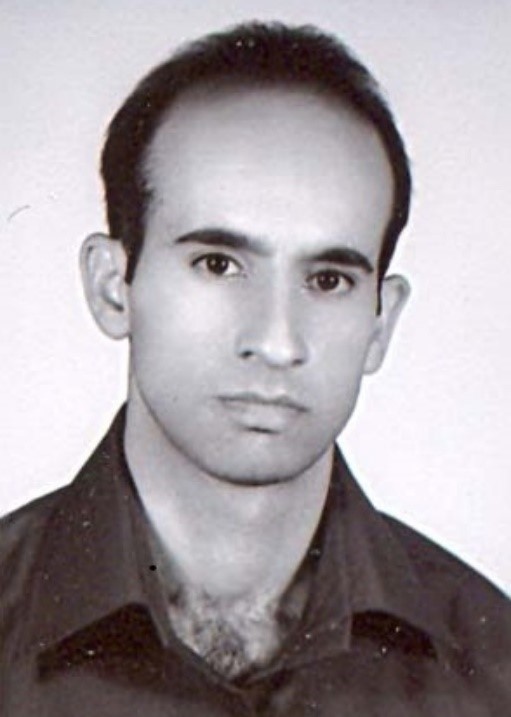}
\end{minipage}
\hfill
\begin{minipage}[t]{0.74\linewidth}
\vspace{0pt}
\textbf{Mohammad Tashakoripour} is a Researcher at the Gastroenterology
Department, Amiralam Hospital, Tehran University of Medical Sciences, Tehran, Iran.
His work focuses on histopathological grading and validation of gastrointestinal
precancerous lesions.
\end{minipage}

\bigskip
\noindent\begin{minipage}[t]{0.22\linewidth}
\vspace{0pt}
\includegraphics[width=\linewidth]{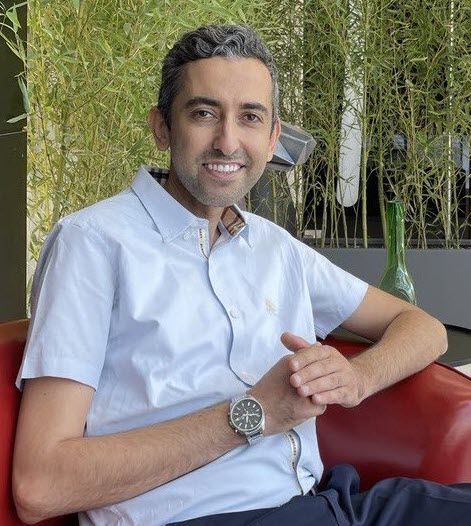}
\end{minipage}
\hfill
\begin{minipage}[t]{0.74\linewidth}
\vspace{0pt}
\textbf{Hamidreza Bolhasani, PhD} is an AI/ML Researcher, Visiting
Professor, Founder and Chief Data Scientist at DataBioX
(\url{https://databiox.com}). He holds a PhD in Computer Engineering from the
Science and Research Branch, Islamic Azad University, Tehran, Iran (2018--2023).
His fields of interest include Artificial Intelligence, Machine Learning, Deep
Learning, Computer Vision, Neural Networks, Computer Architecture, and Bioinformatics.
\end{minipage}
\end{document}